# Optimization under Uncertainty in the Era of Big Data and Deep Learning: When Machine Learning Meets Mathematical Programming

Chao Ning, Fengqi You[*]

Cornell University, Ithaca, New York 14853, USA

## Abstract

This paper reviews recent advances in the field of optimization under uncertainty via a modern data lens, highlights key research challenges and promise of data-driven optimization that organically integrates machine learning and mathematical programming for decision-making under uncertainty, and identifies potential research opportunities. A brief review of classical mathematical programming techniques for hedging against uncertainty is first presented, along with their wide spectrum of applications in Process Systems Engineering. A comprehensive review and classification of the relevant publications on data-driven distributionally robust optimization, data-driven chance constrained program, data-driven robust optimization, and data-driven scenario-based optimization is then presented. This paper also identifies fertile avenues for future research that focuses on a closed-loop data-driven optimization framework, which allows the feedback from mathematical programming to machine learning, as well as scenario-based optimization leveraging the power of deep learning techniques. Perspectives on online learning-based data-driven multistage optimization with a learning-while-optimizing scheme is presented.

[*]Corresponding author. Phone: (607) 255-1162; Fax: (607) 255-9166; E-mail: fengqi.you@cornell.edu

# 1. Introduction

Optimization applications abound in many areas of science and engineering [1-3]. In real practice, some parameters involved in optimization problems are subject to uncertainty due to a variety of reasons, including estimation errors and unexpected disturbance [4]. Such uncertain parameters can be product demands in process planning [5], kinetic constants in reaction-separation-recycling system design [6], and task durations in batch process scheduling [7], among others. The issue of uncertainty could unfortunately render the solution of a deterministic optimization problem (i.e. the one disregarding uncertainty) suboptimal or even infeasible [8]. The infeasibility, i.e. the violation of constraints in optimization problems, has a disastrous consequence on the solution quality. Motivated by the practical concern, optimization under uncertainty has attracted tremendous attention from both academia and industry [4, 9].

In the era of big data and deep learning, intelligent use of data has a great potential to benefit many areas. Although there is no rigorous definition of big data [10], people typically characterize big data with five Vs, namely, volume, velocity, variety, veracity and value [11]. Torrents of data are routinely collected and archived in process industries, and these data are becoming an increasingly important asset in process control, operations and design [12-14]. Nowadays, a wide array of emerging machine learning tools can be leveraged to analyze data and extract accurate, relevant, and useful information to facilitate knowledge discovery and decision-making. Deep learning, one of the most rapidly growing machine learning subfields, demonstrates remarkable power in deciphering multiple layers of representations from raw data without any domain expertise in designing feature extractors [15]. More recently, dramatic progress of mathematical programming [16], coupled with recent advances in machine learning [17], especially in deep learning over the past decade [18], sparks a flurry of interest in data-driven optimization [19-26]. In the data-driven optimization paradigm, uncertainty model is formulated based on data, thus allowing uncertainty data "speak" for themselves in the optimization algorithm. In this way, rich information underlying uncertainty data can be harnessed in an automatic manner for smart and data-driven decision making.

In this review paper, we summarize and classify the existing contributions of data-driven optimization under uncertainty, highlight the current research trends, point out the research challenges, and introduce promising methodologies that can be used to tackle these challenges. We briefly review conventional mathematical programming techniques for hedging against uncertainty, alongside their wide spectrum of applications in Process Systems Engineering (PSE). We then summarize the existing research papers on data-driven optimization under uncertainty and classify them into four categories according to their unique approach for uncertainty modeling and distinct optimization structures. Based on the literature survey, we identify three promising future research directions on optimization under uncertainty in the era of big data and deep learning and highlight respective research challenges and potential methodologies.

The rest of this paper is organized as follows. In Section 2, background on mathematical programming techniques for decision making under uncertainty is given. Section 3 presents a comprehensive literature review, where relevant research papers are summarized and classified into four categories. Section 4 discusses promising future research directions to further advance the area of data-driven optimization. Conclusions are provided in Section 5.

## 2. Background on optimization under uncertainty

In recent years, mathematical programming techniques for decision making under uncertainty have gained tremendous popularity among the PSE community, as witnessed by various successful applications in process synthesis and design [9, 27], production scheduling and planning [7, 28], and process control [29-31]. In this section, we present some background knowledge of methodologies for optimization under uncertainty, along with computational algorithms and applications in PSE. Specifically, we briefly review three leading modeling paradigms for optimization under uncertainty, namely stochastic programming, chance-constrained programming, and robust optimization.

### 2.1. Stochastic programming

Stochastic programming is a powerful modeling paradigm for decision making under uncertainty that aims to optimize the expected objective value across all the uncertainty realizations [32]. The key idea of the stochastic programming approach is to model the randomness in uncertain parameters with probability distributions [33, 34]. In general, the stochastic programming approach can effectively accommodate decision making processes with various time stages. In single-stage stochastic programs, there are no recourse variables and all the decisions must be made before knowing uncertainty realizations. By contrast, stochastic programming with recourse can take corrective actions after uncertainty is revealed. Among the stochastic programming approach with recourse, the most widely used one is the two-stage stochastic program, in which decisions are partitioned into "here-and-now" decisions and "wait-and-see" decisions.

The general mathematical formulation of a two-stage stochastic programming problem is given as follows [32].

$$\min_{\mathbf{x} \in X} \ \mathbf{c}^\mathrm{T} \mathbf{x} + \mathbb{E}_\omega \left[ Q(\mathbf{x}, \omega) \right] \tag{1}$$
$$\text{s.t.} \ \mathbf{A}\mathbf{x} \geq \mathbf{d}$$

The recourse function $Q(\mathbf{x}, \omega)$ is defined by,

$$Q(\mathbf{x}, \omega) = \min_{\mathbf{y}(\omega) \in Y} \ \mathbf{b}(\omega)^\mathrm{T} \mathbf{y}(\omega) \tag{2}$$
$$\text{s.t.} \ \mathbf{W}(\omega) \mathbf{y}(\omega) \geq \mathbf{h}(\omega) - \mathbf{T}(\omega) \mathbf{x}$$

where $\mathbf{x}$ represents first-stage decisions made "here-and-now" before the uncertainty $\omega$ is realized, while the second-stage decisions $\mathbf{y}$ are postponed in a "wait-and-see" manner after observing the uncertainty realization. The objective of the two-stage stochastic programming model includes two parts: the first-stage objective $\mathbf{c}^\mathrm{T}\mathbf{x}$ and the expectation of the second-stage objective $\mathbf{b}(\omega)^\mathrm{T}\mathbf{y}(\omega)$. The constraints associated with the first-stage decisions are $\mathbf{A}\mathbf{x} \geq \mathbf{d}$, $\mathbf{x} \in X$, and the constraints of the second-stage decisions are $\mathbf{W}(\omega)\mathbf{y}(\omega) \geq \mathbf{h}(\omega) - \mathbf{T}(\omega)\mathbf{x}$ and $\mathbf{y}(\omega) \in Y$. Sets $X$ and $Y$ can include nonnegativity, continuity or integrality restrictions.

The resulting two-stage stochastic programming problem is computationally expensive to solve because of the growth of computational time with the number of scenarios. To this end, decomposition based algorithms have been developed in the existing literature, including Benders decomposition or the L-shaped method [35, 36], and Lagrangean decomposition [37]. The location of binary decision variables is critical for the design of computational algorithms. For stochastic programs with integer recourse, the expected recourse function is no longer convex, and even discontinuous, thus hindering the employment of conventional L-shaped method. As a result, research efforts have made on computational algorithms for efficient solution of two-stage stochastic mixed-integer programs [38], such as Lagrangian relaxation [39], branch-and-bound scheme [40], and an improved L-shaped method [41, 42].

Stochastic programming has demonstrated various applications in PSE, such as design and operations of batch processes [43-46], process flowsheet optimization [47], energy systems [48-51], and supply chain management [52-58]. Due to its wide applicability, immense research efforts have been made on the variants of stochastic programming approach. For instance, the two-stage formulation in (1) can be readily extended to a multi-stage stochastic programming setup by utilizing scenario trees. Other extensions include stochastic nonlinear programming [59], and stochastic programs with endogenous uncertainties [60, 61].

## 2.2. Chance constrained optimization

As another powerful paradigm for optimization under uncertainty, chance constrained programming aims to optimize an objective while ensuring constraints to be satisfied with a specified probability in uncertain environment [62]{Uryasev, 2000 #2757}. As in the stochastic programming approach, probability distribution is the key uncertainty model to capture the randomness of uncertain parameters in chance constrained optimization. The chance constrained program was first introduced in the seminal work of [63], and attracted considerable attention ever since. Such chance constraints or probabilistic constraints are flexible enough to quantify the trade-off between objective performance and system reliability [64].

The generic formulation of a chance constrained optimization problem is presented as follows,

$$\min_{\mathbf{x} \in X} f(\mathbf{x})$$
$$\text{s.t. } \mathbb{P}\{\boldsymbol{\xi} \in \Xi \mid G(\mathbf{x}, \boldsymbol{\xi}) \leq \mathbf{0}\} \geq 1 - \varepsilon \quad (3)$$

where $\mathbf{x}$ represents the vector of decision variables, $X$ denotes the deterministic feasible region, $f$ is the objective function to be minimized, $\boldsymbol{\xi}$ is a random vector following a known probability distribution $\mathbb{P}$ with the support set $\Xi$, $G = (g_1, \ldots, g_m)$ represents a constraint mapping, $\mathbf{0}$ is a vector of all zeros, and parameter $\varepsilon$ is a pre-specified risk level.

The chance constraint $\mathbb{P}\{\boldsymbol{\xi} \in \Xi \mid G(\mathbf{x}, \boldsymbol{\xi}) \leq \mathbf{0}\} \geq 1 - \varepsilon$ guarantees that decision $\mathbf{x}$ satisfies constraints with a probability of at least $1-\varepsilon$. Note that when the number of constraints $m=1$, the above optimization model is an individual chance constrained program; for $m>1$, it is called joint chance constrained program [65]. A salient merit of chance constrained programs is that it allows decision makers choose their own risk levels for the improvement in objectives. To model sequential decision-making processes, two-stage chance constrained optimization with recourse was recently studied and had various applications [66, 67].

Despite of its promising modeling power, the resulting chance constrained program is generally computationally intractable for the following two main reasons. First, calculating the probability of constraint satisfaction for a given x involves a multivariate integral, which is believed to be computationally prohibitive. Second, the feasible region is not convex even if set *X* is convex and *G*(**x**, ξ) is convex in **x** for any realizations of uncertain vector ξ [62]. In light of these computational challenges, a large body of related literature is devoted into the development of solution algorithms for chance constrained optimization problems, such as sample average approximation [68], sequential approximation [69], and convex conservative approximation schemes [70]. Note that chance constrained programs admit convex reformulation for some very special cases. For example, individual chance constrained programs are endowed with tractable convex reformulations for normal distributions [32]. Chance constraints with right-hand-side uncertainty are convex if uncertain parameters are independent and follow log-concave distributions [62].

In the PSE community, chance constraints are usually employed for customer demand satisfaction, product quality specification, and service level of process systems {Maranas, 1997 #745;Yue, 2013 #2254;Gupta, 2000 #2377;You, 2011 #2278;Chu, 2015 #3322;Zipkin, 2000 #2521}. Due to its practical relevance, chance constrained optimization has been applied in numerous applications, including model predictive control [76, 77], process design and operations [78-80], refinery blend planning [81], biopharmaceutical manufacturing [82], and supply chain optimization [83-86].

## 2.3. Robust optimization

As a promising alternative paradigm, robust optimization does not require accurate knowledge on probability distributions of uncertain parameters [87-89]. Instead, it models uncertain parameters using an uncertainty set, which includes possible uncertainty realizations. It is worth noting that uncertainty set is a paramount ingredient in robust optimization framework [89]. Given a specific uncertainty set, the idea of robust optimization is to hedge against the worst case within the uncertainty set. The worst-case uncertainty realization is defined based on different contexts: it could be the realization giving rise to the largest constraint violation, the realization leading to the lowest asset return [90] or the one resulting in the highest regret [91].

The conventional box uncertainty set is not a good choice since it includes the unlikely-to-happen scenario where uncertain parameters simultaneously increase to their highest values. The conventional box uncertainty set is defined as follows [92].

$$U_{\text{box}} = \left\{ \mathbf{u} \mid \mathbf{u}_i^L \leq \mathbf{u}_i \leq \mathbf{u}_i^U, \ \forall i \right\} \tag{4}$$

where $U_{\text{box}}$ is a box uncertainty set, $\mathbf{u}$ is a vector of uncertain parameters, $\mathbf{u}_i$ is the $i$-th component of uncertainty vector $\mathbf{u}$. $\mathbf{u}_i^L$ and $\mathbf{u}_i^U$ represent the lower bound and the upper bound of uncertain parameter $\mathbf{u}_i$, respectively. Box uncertainty set simply defines the range of each uncertain parameter in vector $\mathbf{u}$. One cannot easily control the size of this uncertainty set to meet his or her risk-averse attitude. To this end, researchers propose the following budgeted uncertainty set [88].

$$U_{\text{budget}} = \left\{ \mathbf{u} \mid \mathbf{u}_i = \overline{\mathbf{u}}_i + \Delta \mathbf{u}_i \cdot z_i, \ -1 \leq z_i \leq 1, \ \sum_i |z_i| \leq \Gamma, \ \forall i \right\} \tag{5}$$

where $U_{\text{budget}}$ denotes a budgeted uncertainty set, $\mathbf{u}$ and $\mathbf{u}_i$ have the same definitions as in (4), $\overline{\mathbf{u}}_i$ is the nominal value of $\mathbf{u}_i$, $\Delta \mathbf{u}_i$ is the largest possible deviation of uncertain parameter $\mathbf{u}_i$, $z_i$ denotes the extent and direction of parameter deviation, and $\Gamma$ is an uncertainty budget.

Traditional robust optimization approaches, also known as static robust optimization [93], make all the decisions at once. This modeling framework cannot well represent sequential decision-making problems [94-96]. Adaptive robust optimization (ARO) was proposed to offer a new paradigm for optimization under uncertainty by incorporating recourse decisions [97]. Due to the flexibility of adjusting recourse decisions after observing uncertainty realizations, ARO typically generates less conservative solutions than static robust optimization [98]. The general form of a two-stage adaptive robust mixed-integer programming model is given as follows:

$$\begin{aligned}
& \min_{\mathbf{x}} \mathbf{c}^T \mathbf{x} + \max_{\mathbf{u} \in U} \min_{\mathbf{y} \in \Omega(\mathbf{x}, \mathbf{u})} \mathbf{b}^T \mathbf{y} \\
& \text{s.t.} \quad \mathbf{A}\mathbf{x} \geq \mathbf{d}, \quad \mathbf{x} \in R_+^{n_1} \times Z^{n_2} \\
& \quad \Omega(\mathbf{x}, \mathbf{u}) = \left\{ \mathbf{y} \in R_+^{n_3} : \mathbf{W}\mathbf{y} \geq \mathbf{h} - \mathbf{T}\mathbf{x} - \mathbf{M}\mathbf{u} \right\}
\end{aligned} \tag{6}$$

where $\mathbf{x}$ is the first-stage decision made before uncertainty $\mathbf{u}$ is realized, while the second-stage decision $\mathbf{y}$ is postponed in a "wait-and-see" manner. $\mathbf{x}$ includes both continuous and integer variables, while $\mathbf{y}$ only includes continuous variables. $\mathbf{c}$ and $\mathbf{b}$ are the vectors of the cost coefficients. $U$ is an uncertainty set that characterizes the region of uncertainty realizations. ARO approaches could be applied to address uncertainty in a variety of applications, including process design [99-101], process scheduling [102], supply chain optimization [101, 103], among others. Besides the two-stage ARO framework, the multistage ARO method has attracted immense attention due to its unique feature in reflecting sequential realizations of uncertainties over time [104]. In multistage ARO, decisions are made sequentially, and uncertainties are revealed gradually over stages. Note that the additional value delivered by ARO over static robust optimization is its adjustability of recourse decisions based on uncertainty realizations [97]. Accordingly, the multistage ARO method has demonstrated applications in process scheduling and planning [105, 106].

Despite popularity of the above three leading paradigms for optimization under uncertainty, these approaches have their own limitations and specific application scopes. To this end, research efforts have been made on "hybrid" methods that leverage the synergy of different optimization approaches to inherit their corresponding strengths and complement respective weaknesses [107-112]. For instance, stochastic programming was integrated with robust optimization for supply chain design and operation under multi-scale uncertainties [50]. Robust chance constrained optimization along with global solution algorithms were developed and applied to process design under price and demand uncertainties [112].

## 3. Existing methods for data-driven optimization under uncertainty

In this section, we review the recent advances in optimization under uncertainty in the era of big data and deep learning. Recent years have witnessed a rapidly growing number of publications on data-driven optimization under uncertainty, an active area integrating machine learning and mathematical programming. These publications cover various topics and can be roughly classified into four categories, namely data-driven stochastic program, data-driven chance constrained program, data-driven robust optimization, and data-driven scenario-based optimization. Unlike the conventional mathematical programming techniques, these data-driven approaches do not presume the uncertainty model is perfectly given *a priori*, rather they all focus on the practical setting where only uncertainty data are available.

### 3.1. Data-driven stochastic program and distributionally robust optimization

The literature review of data-driven stochastic program, also known as distributionally robust optimization (DRO), is presented in detail in this subsection. The motivation of this emerging paradigm on data-driven optimization under uncertainty is first presented, followed by its model formulation. In this modeling paradigm, the uncertainty is modeled via a family of probability distributions that well capture uncertainty data on hand. This set of probability distributions is referred to as ambiguity set. We then present and analyze various types of ambiguity sets alongside their corresponding strengths and weaknesses. Finally, the extension of DRO to the multistage decision-making setting is also discussed, as well as their recent applications in PSE.

In the stochastic programming approach, it is assumed that the probability distribution of uncertain parameters is perfectly known. However, such precise information of the uncertainty distribution is rarely available in practice. Instead, what the decision maker has is a set of historical and/or real-time uncertainty data and possibly some prior structure knowledge of the probability. Moreover, the assumed probability in conventional stochastic programming might deviate from the true distribution. Therefore, relying on a single probability distribution could result in sub-optimal solutions, or even lead to the deterioration in out-of-sample performance [113]. Motivated by these weaknesses of stochastic programming, DRO emerges as a new data-driven optimization paradigm which hedges against the worst-case distribution in an ambiguity set. Rather than assuming a single uncertainty distribution, the DRO approach constructs an uncertainty set of probability distributions from uncertainty data through statistical inference and big data analytics. In this way, DRO is capable of hedging against the distribution errors, and accounts for the input of uncertainty data.

The general model formulation of data-driven stochastic programming is presented as follows [114].

$$\min_{\mathbf{x}\in X} \max_{\mathbb{P}\in\mathcal{D}} \mathbb{E}_{\mathbb{P}}\left\{l\left(\mathbf{x},\xi\right)\right\} \tag{7}$$

where $\mathbf{x}$ is the vector of decision variables, $X$ is the feasible set, $l$ is the objective function, and $\xi$ represents a random vector whose probability distribution $\mathbb{P}$ is only known to reside in an ambiguity set $\mathcal{D}$. The DRO approach aims for optimal decisions under the worst-case distribution, and as a result offers performance guarantee over the family of distributions.

The DRO or data-driven stochastic optimization framework enjoys two salient merits compared with the conventional stochastic programming approach. First, it allows the decision maker to incorporate partial distribution information learned from uncertainty data into the optimization. As a result, the data-driven stochastic programming approach greatly mitigates the issue of optimizer's curse and improves the out-of-sample performance. Second, data-driven stochastic programming inherits the computational tractability from robust optimization and some resulting problems can be solved exactly in polynomial time without resorting to the approximation scheme via sampling or discretization. For example, optimization problem (7) for a convex program with continuous variables and a moment-based ambiguity set is proved to be solvable in polynomial time [114].

The choice of ambiguity sets plays a critical role in the performance of DRO. When choosing ambiguity set, the decision maker need to consider the following three factors, namely tractability, statistical meaning, and performance [115]. First, the data-driven stochastic programming problem with the ambiguity set should be computationally tractable, meaning the resulting optimization could be formulated as linear, conic quadratic or semidefinite programs. Second, the derived ambiguity set should have clear statistical meaning. Therefore, various ways of constructing ambiguity sets based on uncertainty data were extensively studied [114, 116, 117]. Third, the devised ambiguity set should be tight to increase the performance of resulting decisions.

One commonly used approach to constructing ambiguity set is moment-based approaches, in which first and second order information is extracted from uncertainty data using statistical inference [118]. The ambiguity set that specifies the support, first and second moment information is shown as follows,

$$\mathcal{D} = \left\{ \mathbb{P} \in \mathcal{M}_+ \;\middle|\; \begin{array}{l} \mathbb{P}[\xi \in \Xi] = 1 \\ \mathbb{E}_\mathbb{P}[\xi] = \mu \\ \mathbb{E}_\mathbb{P}\left[(\xi - \mu)(\xi - \mu)^T\right] = \Sigma \end{array} \right\} \quad (8)$$

where $\xi$ represents the uncertainty vector, $\Xi$ is the support, $\mathbb{P}$ represents the probability distribution of $\xi$, $\mathcal{M}_+$ denotes the set of all probability measures, $\mathbb{E}_\mathbb{P}$ denotes the expectation with respect to distribution $\mathbb{P}$. Parameters $\mu$ and $\Sigma$ represent the mean vector and covariance matrix estimated from uncertainty data, respectively.

The ambiguity set in (8) fails to account for the fact that the mean and covariance matrix are also subject to uncertainty. To this end, an ambiguity set was proposed based on the distribution's support information as well as the confidence regions for the mean and second-moment matrix in the work of [114]. The resulting DRO problem could be solved efficiently in polynomial time.

$$\mathcal{D} = \left\{ \mathbb{P} \in \mathcal{M}_+ \;\middle|\; \begin{array}{l} \mathbb{P}[\xi \in \Xi] = 1 \\ \left(\mathbb{E}_\mathbb{P}[\xi] - \mu\right)^T \Sigma^{-1} \left(\mathbb{E}_\mathbb{P}[\xi] - \mu\right) \leq \psi_1 \\ \mathbb{E}_\mathbb{P}\left[(\xi - \mu)(\xi - \mu)^T\right] \leq \psi_2 \Sigma \end{array} \right\} \quad (9)$$

where $\xi$ represents the uncertainty vector, $\Xi$ is the support, $\mathbb{P}$ represents the probability distribution of $\xi$. The equality constraint $\mathbb{P}[\xi \in \Xi]=1$ enforces that all uncertainty realizations reside in the support set $\Xi$. Parameters $\psi_1$ and $\psi_2$ are used to define the sizes of confidence regions for the first and second moment information, respectively.

The moment-based ambiguity sets typically enjoy the advantage of computational tractability. For example, DRO with the ambiguity set based on principal component analysis and first-order deviation functions was developed [117]. Additionally, the computational effectiveness of this data-driven DRO method was demonstrated via process network planning and batch production scheduling . Recently, a data-driven DRO model was developed for the optimal design and operations of shale gas supply chains to hedge against uncertainties associated with shale well estimated ultimate recovery and product demand [119]. However, the moment-based ambiguity set is not guaranteed to converge to the true probability distribution as the number of uncertainty data goes to infinity. Consequently, this type of ambiguity set suffers from the conservatism with moderate uncertainty data. To address the above issue with moment-based methods, ambiguity sets based on statistical distance between probability distributions were developed, as shown below,

$$\mathcal{D} = \left\{ \mathbb{P} \in \mathcal{M}_+ \mid d(\mathbb{P}, \mathbb{P}_0) \leq \theta \right\} \tag{10}$$

where $\mathbb{P}$ is the probability distribution of uncertain parameters, $\mathbb{P}_0$ represents the reference distribution such as the empirical distribution, $d$ denotes some statistical distance between two distributions, and $\theta$ stands for the confidence level.

Ambiguity set in (10) can be further classified based on the adopted distance metric, such as Kullback-Leibler divergence [120] and Wasserstein distance [116]. For example, a DRO model was proposed for lot-sizing problem, in which the chi-square goodness-of-fit test and robust optimization were combined. The ambiguity set of demand was constructed from uncertainty data by using a hypothesis test in statistics, called the chi-square goodness-of-fit test [121]. This set is well defined by linear constraints and second order cone constraints. It is worth noting that the input of their model is histograms, which make it possible to use a finite dimensional probability vector to characterize the distribution. The adopted statistic belonged to the phi-divergences, which motivated researchers to construct distribution uncertainty set by using the phi-divergences [122].

To account for the sequential decision-making process, researchers recently developed the adaptive DRO method by incorporating recourse decision variables [123, 124]. A general two-stage data-driven stochastic programming model is presented in the following form:

$$\min_{\mathbf{x} \in X} \mathbf{c}^T \mathbf{x} + \max_{\mathbb{P} \in \mathcal{D}} \mathbb{E}_{\mathbb{P}} \{Q(\mathbf{x}, \xi)\}$$
$$\text{s.t.} \quad \mathbf{A}\mathbf{x} \leq \mathbf{d} \tag{11}$$
$$Q(\mathbf{x}, \xi) = \begin{cases} \min_{\mathbf{y} \in Y} \mathbf{b}(\xi)^T \mathbf{y} \\ \text{s.t.} \quad \mathbf{T}(\xi)\mathbf{x} + \mathbf{W}(\xi)\mathbf{y} \leq \mathbf{h}(\xi) \end{cases}$$

where **x** presents the vector of first-stage decision variables that need to be determined before observing uncertainty realizations, **y** denotes the vector of second-stage decision variables that can be adjustable based on the realized uncertain parameters $\xi$, sets $X$ and $Y$ can include nonnegativity, continuity or integrality restrictions, and $Q$ represents the recourse function. The objective of the above data-driven stochastic program is to minimize the worst-case expected cost with respect to all possible uncertainty distributions $\mathbb{P}$ within the ambiguity set $\mathcal{D}$. Based on the literature, multistage data-driven DRO is becoming a rapidly evolving research direction.

Data-driven stochastic programming has several salient merits over the conventional stochastic programming approach. However, there are few papers on its PSE applications in the existing literature [117, 119]. As the trend of big data has fueled the increasing popularity of data-driven stochastic programming in many areas, DRO emerges as a new data-driven optimization paradigm which hedges against the worst-case distribution in an ambiguity set, and has various applications in power systems, such as unit commitment problems [125-128], and optimal power flow [129, 130].

## 3.2. Data-driven chance constrained program

In contrast to the data-driven stochastic programming approach reviewed in Section 3.1, data-driven chance constrained programming is another paradigm focusing on chance constraint satisfaction under the worst-case probability instead of optimizing the worst-case expected objective. Although both data-driven chance constrained program and DRO adopt ambiguity sets in the uncertainty models, they have distinct model structures. Specifically, data-driven chance constrained program features constraints subject to uncertainty in probability distributions, while DRO typically only involves the worst-case expectation of an objective function with respect to a family of probability distributions. As introduced in Section 2, the chance constrained programming approach assumes the complete distribution information is perfectly known. However, the decision maker only has access to a finite number of uncertainty realizations or uncertainty data. On one hand, such complete knowledge of distribution is usually estimated from limited number of uncertainty data or obtained from expert knowledge. On the other hand, even if the probability distribution is available, the chance constrained program is computationally cumbersome. In practice, one can only have partial information on the probability distribution of uncertainty. Therefore, data-driven chance constrained optimization emerges as another paradigm for hedging against uncertainty in the era of big data.

The general form of data-driven chance constrained program is given by,

$$\min_{\mathbf{x} \in X} f(\mathbf{x})$$
$$\text{s.t.} \quad \min_{\mathbb{P} \in \mathcal{D}} \mathbb{P}\{\boldsymbol{\xi} \in \Xi \mid G(\mathbf{x}, \boldsymbol{\xi}) \leq \mathbf{0}\} \geq 1 - \varepsilon \quad (12)$$

where $\mathbf{x}$ represents the vector of decision variables, $X$ denotes the deterministic feasible region, $f$ is the objective function, $\boldsymbol{\xi}$ is a random vector following a probability distribution $\mathbb{P}$ that belongs to an ambiguity set $\mathcal{D}$. $G = (g_1, \ldots, g_m)$ represents a constraint mapping, $\mathbf{0}$ is a vector of all zeros, and parameter $\varepsilon$ is a pre-specified risk level. The data-driven chance constraints enforce classical chance constraints to be satisfied for every probability distribution within the ambiguity set.

The computational tractability of the resulting data-driven chance constrained program can vary depending on both the ambiguity sets and the structure of the optimization problem. In the following, we summarize the relevant papers according to the adopted uncertainty set of distributions and optimization structures.

Distributionally robust individual linear chance constraints under the ambiguity set comprised of all distributions sharing the same known mean and covariance were reformulated as convex second-order cone constraints [118]. The deterministic convex conditions to enforce distributionally robust chance constraints were provided under distribution families of (a) independent random variables with box-type support and (b) radially symmetric non-increasing distributions over the orthotope support. The worst-case conditional value-at-risk (CVaR) approximation for distributionally robust joint chance constraints was studied assuming first and second moment [131], and the resulting conservative approximation can be cast as semidefinite program. In addition to moment information, a specific structural information of distributions called unimodality was incorporated into the ambiguity set, and the corresponding ambiguous risk constraints were reformulated as a set of second second-order cone constraints [132]. Instead of assuming unimodality of distributions, data-driven robust individual chance constrained programs along with convex approximations were recently developed using a mixture distribution-based ambiguity set with fixed component distribution and uncertain mixture weights [133].

In real world applications, exact moment information can be challenging to obtain, and can only be estimated through confidence intervals from uncertainty realizations [114]. To accommodate this moment uncertainty, attempts were made in the context of distributionally robust chance constraints, including constructing convex moment ambiguity set [134], employing Chebyshev ambiguity set with bounds on second-order moment [135], characterizing a family of distributions with upper bounds on both mean and covariance [136]. Ambiguous joint chance constraints were studied where the ambiguity set was characterized by the mean, convex support, and an upper bound on the dispersion [137], and the resulting constraints were conic representable for right-hand-side uncertainty. In addition to generalized moment bounds [138], structural properties of distributions, such as symmetry, unimodality, multimodality and independence, were further integrated into distributionally robust chance constrained programs leveraging a Choquet representation [115]. Nonlinear extensions of distributionally robust chance constraints were made under the ambiguity sets defined by mean and variance [139], convex moment constraints [140], mean absolute deviation [141], and a mixture of distributions [142].

Although moment-based ambiguity sets achieve certain success, they do not converge to the true probability distribution as the number of available uncertainty data increases. Consequently, the resulting data-driven chance-constrained programs tend to generate conservative solutions. To this end, data-driven chance-constrained programs with distance-based ambiguity set were proposed to alleviate the undesirable consequence of moment-based data-driven chance-constrained programs. The ambiguity set defined by the Prohorov metric was introduced into the distributionally robust chance constraints, and the resulting optimization problem was approximated by using robust sampled problem [143]. Distributionally robust chance constraints with the ambiguity set containing all distributions close to a reference distribution in terms of Kullback-Leibler divergence were cast as classical chance constraints with an adjusted risk level [120]. Data-driven chance constrained programs with $\phi$-divergence based ambiguity set were proposed [144], and further extensions were made using the kernel smoothing method [22].

Recently, data-driven chance constraints over Wasserstein balls were exactly reformulated as mixed-integer conic constraints [145, 146]. Leveraging the strong duality result [147], distributionally robust chance constrained programs with Wasserstein ambiguity set were studied for linear constraints with both right and left hand uncertainty [148], as well as for general nonlinear constraints [149].

Data-driven chance constrained programs have successful applications in a number of areas, such as power system [150], stochastic control [151], and vehicle routing problem [152].

### 3.3. Data-driven robust optimization

As a paramount ingredient in robust optimization, uncertainty sets endogenously determine robust optimal solutions and therefore should be devised with special care. However, uncertainty sets in the conventional robust optimization methodology are typically set *a priori* using a fixed shape and/or model without providing sufficient flexibility to capture the structure and complexity of uncertainty data. For example, the geometric shapes of uncertainty sets in (4) and (5) do not change with the intrinsic structure and complexity of uncertainty data. Furthermore, these uncertainty sets are specified by finite number of parameters, thereby having limited modeling flexibility. Motivated by this knowledge gap, data-driven robust optimization emerges as a powerful paradigm for addressing uncertainty in decision making.

A data-driven ARO framework that leverages the power of Dirichlet process mixture model was proposed [25]. The data-driven approach for defining uncertainty set was developed based on Bayesian machine learning. This machine learning model was then integrated with the ARO method through a four-level optimization framework. This developed framework effectively accounted for the correlation, asymmetry and multimode of uncertainty data, so it generated less conservative solutions. Its salient feature is that multiple basic uncertainty sets are used to provide a high-fidelity description of uncertainties. Although the data-driven ARO has a number of attractive features, it does not account for an important evaluation metric, known as regret, in decision-making [153]. Motivated by the knowledge gap, a data-driven bi-criterion ARO framework was developed that effectively accounted for the conventional robustness as well as minimax regret [154].

In some applications, uncertainty data in large datasets are usually collected under multiple conditions. A data-driven stochastic robust optimization framework was proposed for optimization under uncertainty leveraging labeled multi-class uncertainty data [155]. Machine learning methods including Dirichlet process mixture model and maximum likelihood estimation were employed for uncertainty modeling, which is illustrated in Figure 1. This framework was further proposed based on the data-driven uncertainty model through a bi-level optimization structure. The outer optimization problem followed the two-stage stochastic programming approach, while ARO was nested as the inner problem for maintaining computational tractability.

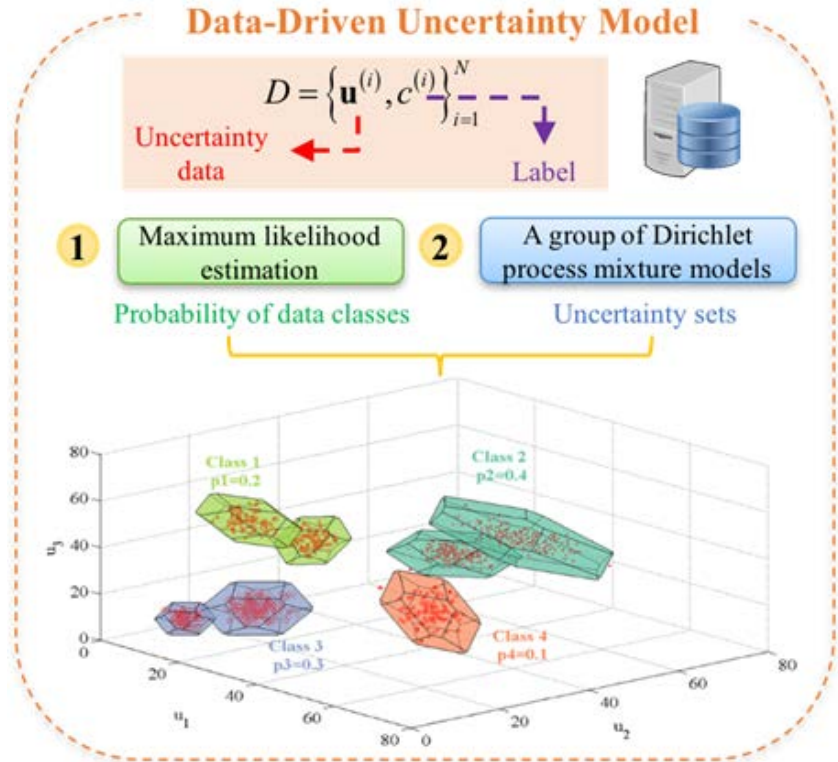

**Figure 1.** The data-driven uncertainty model based on the Dirichlet process mixture model [155].

To mitigate computational burden, research effort has been made on convex polyhedral data-driven uncertainty set based on machine learning techniques, such as principal component analysis and support vector clustering. A data-driven robust optimization framework that leveraged the power of principal component analysis and kernel smoothing for decision-making under uncertainty was studied [156]. In this approach, correlations between uncertain parameters were effectively captured, and latent uncertainty sources were identified by principal component analysis. To account for asymmetric distributions, forward and backward deviation vectors were utilized in the uncertainty set, which was further integrated with robust optimization models. A data-driven static robust optimization framework based on support vector clustering that aims to find the hypersphere with minimal volume to enclose uncertainty data was proposed [157]. The adopted piecewise linear kernel incorporates the covariance information, thus effectively capturing the correlation among uncertainties. These two data-driven robust optimization approaches utilized polyhedral uncertainty learned from data, and thus enjoying computational efficiency. Various types of data-driven uncertainty sets were developed for static robust optimization based on statistical hypothesis tests [19], copula [158], and probability density contours [159].

To address multistage decision making under uncertainty, a data-driven approach for optimization under uncertainty based on multistage ARO and nonparametric kernel density M-estimation was developed [106]. The salient feature of the framework was its incorporation of distributional information to address the issue of over-conservatism. Robust kernel density estimation was employed to extract probability distributions from data. This data-driven multistage ARO framework exploited robust statistics to be immunized to data outliers. An exact robust counterpart was developed for solving the resulting data-driven ARO problem.

In recent years, data-driven robust optimization has been applied to a variety of areas, such as power systems [160], energy systems [161], planning and scheduling [106, 158], process control [156, 162], and transportation systems [163, 164].

### 3.4. Scenario optimization approach for chance constrained programs

A salient feature of scenario-based optimization is that it does not require the explicit knowledge of probability distribution as in the stochastic programming approach. Additionally, scenario-based optimization uses uncertainty scenarios to seek an optimal solution having a high probabilistic guarantee of constraint satisfaction instead of utilizing scenarios or samples to approximate the expectation term as in stochastic programming. Although the scenario-based optimization can be regarded as a special type of robust optimization that has a discrete uncertainty set consisting of uncertainty data, it can provide probabilistic guarantee for those unobserved uncertainty data in the testing data set. Note that the scenario-based optimization approach provides a viable and data-driven route to achieving approximate solutions of chance-constrained programs. The scenario-based optimization approach is a general data-driven optimization under uncertainty framework in which uncertainty data or random samples are utilized in a more direct manner compared with other data-driven optimization methods. This data-driven optimization framework was first introduced in [165], and has gained great popularity within the systems and control community [166]. As in data-driven chance constrained programs, the knowledge of true underlying uncertainty distribution is not required in scenario optimization but a finite number of uncertainty realizations. Specifically, the scenario approach enforces the constraint satisfaction with $N$ independent identically distributed uncertainty data $\mathbf{u}^{(1)}$, …, $\mathbf{u}^{(N)}$. The resulting scenario optimization problem is given by,

$$\min_{\mathbf{x} \in X} \quad \mathbf{c}^T \mathbf{x}$$
$$\text{s.t.} \quad f\left(\mathbf{x}, \mathbf{u}^{(i)}\right) \leq 0, \quad i = 1, \ldots, N \tag{13}$$

where $\mathbf{x}$ is the vector of decision variables, $X$ represents a deterministic convex and closed set unaffected by uncertainty, $\mathbf{c}$ is the vector of cost coefficients, and $f$ denotes the constraint function affected by uncertainty $\mathbf{u}$. Note that function $f$ is typically assumed to be convex in $\mathbf{x}$, and can have arbitrarily nonlinear dependence on $\mathbf{u}$, as opposed to data-driven nonlinear chance constrained program assuming the constraint function must be quasi-convex in $\mathbf{u}$ [139]. Additionally, scenario-based optimization can be considered as a special case of data-driven robust optimization when the uncertainty set is constructed as a union of $\mathbf{u}^{(1)}$, …, $\mathbf{u}^{(N)}$.

In the scenario optimization literature, $\omega \doteq \left\{\mathbf{u}^{(1)}, \ldots, \mathbf{u}^{(N)}\right\}$ is referred to as the multi-sample or scenario that is drawn from the product probability space. Due to the random nature of the multi-sample, the optimal solution of the scenario optimization problem (13), denoted as $\mathbf{x}^*(\omega)$, is also random. One key merit of the scenario approach is that the scenario optimization problem admits the same problem type as its deterministic counterpart, so that it can be solved efficiently by convex optimization algorithms when $f(\mathbf{x}, \mathbf{u})$ is convex in $\mathbf{x}$ [167]. Moreover, the optimal solution $\mathbf{x}^*(\omega)$ is guaranteed to satisfy the constraints with other unseen uncertainty realizations with a high probability [168].

For the sake of clarity, we revisit the following definition and theorem [168].

**Definition** (Violation probability) The violation probability of a given decision $\mathbf{x}$ is defined as follows:

$$V(\mathbf{x}) \doteq \mathbb{P}\left\{\mathbf{u} \in \Xi \mid f(\mathbf{x}, \mathbf{u}) > 0\right\} \tag{14}$$

where $V(\mathbf{x})$ denotes the probability of violation for a given $\mathbf{x}$, and $\Xi$ represents the support of uncertainty $\mathbf{u}$. We say a decision $\mathbf{x}$ is $\varepsilon$-feasible if $V(\mathbf{x}) \leq \varepsilon$.

**Theorem**. Assuming $\mathbf{x}^*(\omega)$ is the unique optimal solution of the scenario optimization problem. It holds that

$$\mathbb{P}^N\left\{\omega \big| V\left(\mathbf{x}^*(\omega)\right) \leq \varepsilon\right\} \geq 1 - \sum_{i=0}^{n-1} \binom{N}{i} \varepsilon^i (1-\varepsilon)^{N-i} \tag{15}$$

where $n$ is the number of decision variables, $N$ denotes the number of uncertainty data, and $\mathbb{P}^N$ is a product probability governing the sample generation.

The above theorem implies that the optimal solution $\mathbf{x}^*(\omega)$ satisfies the corresponding chance constraint with a certain confidence level. The proof of this theorem depends on the fundamental fact that the number of support constraints, the removal of which changes the optimal solution, is upper bounded by the number of decision variables [165]. Note that (15) holds with equality for the fully-supported convex optimization problem [168], meaning that the probability bound is tight. Additionally, the result holds true irrespective of probability distribution information or even its support set.

By exploiting the structured dependence on uncertainty, the sample size required by the scenario optimization problem was reduced through a tighter bound on Helly's dimension [169]. Rather than focusing on the constraint violation probability, considerable research efforts have been made on the degree of violation [170], expected probability of constraint violation [171], and the performance bounds for objective values [172]. To make a trade-off between feasibility and performance, the case was studied where some of the sampled constraints were allowed to be violated for improving the performance of the objective [173]. Subsequent work along this direction includes a sampling-and-discarding method [174]. A wait-and-judge scenario optimization framework was proposed in which the level of robustness was assessed *a posteriori* after the optimal solution was obtained [175]. Recently, the extension of scenario-based optimization to the multistage decision making setting was made [176, 177].

While the scenario optimization problems with continuous decision variables are extensively studied [166], the mixed-integer scenario optimization was less developed. An attempt to extend the scenario theory to random convex programs with mixed-integer decision variables was made [178], and the Helly dimension in the mixed-integer scenario program was proved to depend geometrically on the number of integer variables. This result suggests that the required sample size can be prohibitively large for scenario programs with many discrete variables. Along this research direction, two sampling algorithms within the framework of *S*-optimization were recently developed for solving mixed-integer convex scenario programs [179].

In some real-world applications, the required sample size can be very large, resulting in great computational burden for scenario optimization problems with huge number of sampling constraints. One way to circumvent this difficulty is to devise sequential solution algorithms. Along this direction, sequential randomized algorithms were developed for convex scenario optimization problems [180], and fell into the framework of Sequential Probabilistic Validation (SPV) [181]. The motivation behind these sequential algorithms is that validating a given solution with a large number of samples is less computational expensive than solving the corresponding scenario optimization problem. Recently, a repetitive scenario design approach was proposed by iterating between reduced-size scenario optimization problems and the probabilistic feasibility check [182]. The trade-off between the sample size and the expected number of repetitions was also revealed in the repetitive scenario design [182]. Note that the classical scenario-based approach is an extreme situation in the trade-off curve, where one seeks to find the solution at one step. Another effective way to reduce the computation cost of large-scale scenario optimization is to employ distributed algorithms [183, 184]. Particularly, the sampled constraints were distributed among multiple processors of a network, and the large-scale scenario optimization problems can be efficiently solved via constraint consensus schemes [184]. Along this direction, a distributed computing framework was developed for the scenario convex program with multiple processors connected by a graph [185]. The major advantage of this approach is that the computational cost for each processor becomes lower and the original scenario optimization problem can be solved collaboratively. Other contribution to reduce computational cost is made based on a non-iterative two-step procedure, i.e. the optimization step and detuning step [186]. As a consequence, the total sample complexity was greatly decreased.

Traditionally, the field of scenario optimization has focused on convex optimization problems, in which the number of support constraints is upper bounded by the number of decision variables. However, such upper bounds are no longer available in nonconvex scenario optimization problems, giving rise to research challenges of extending the scenario theory to the nonconvex setting. To date, few works have considered nonconvex uncertain program using the scenario approach. One contribution is that of [187], in which assessing the generalization of the optimal solution in a wait-and-judge manner through the concept of support sub-sample was proposed. The proposed approach can be employed to general nonconvex setups, including mixed-integer scenario optimization problems. Another attempt to address nonconvex scenario optimization made use of the statistical learning theory for bounding the violation probability, and devised a randomized solution algorithm [188]. The statistical learning theory-based method provided the probabilistic guarantee for all feasible solutions, as opposed to the convex scenario approach where such guarantee is valid only for the optimal solution. This unique feature regarding probabilistic guarantees for all feasible solutions granted by the statistical learning based method is of practical relevance [189], since it is computationally challenging to solve nonconvex optimization problems to global optimality. A class of non-convex scenario optimization problem, which has non-convex objective functions and convex constraints, was recently studied [190]. Since the Helly's dimension for the optimal solution of such non-convex scenario program can be unbounded, the direct application of scenario approaches based on Helly's theorem is impossible. To overcome the research challenge, the feasible region was restricted to the convex hull of few optimizers, thus enabling the application of sample complexity results [168].

## 4. Future research directions and opportunities

Several promising research directions in data-driven optimization under uncertainty are highlighted in this section. We specifically focus on some ideas on closed-loop data-driven optimization, integration of deep learning and scenario-based optimization, and learning-while-optimizing frameworks.

## 4.1. A "closed-loop" data-driven optimization framework with feedback from mathematical programming to machine learning

The framework of data-driven optimization under uncertainty could be considered as a "hybrid" system that integrates the data-driven system based on machine learning to extract useful and relevant information from data, and the model-based system based on the mathematical programming to derive the optimal decisions from the information. Existing data-driven optimization approaches adopt a sequential and open-loop scheme, which could be further improved by introducing feedback steps from the model-based system to data-driven system. A "closed-loop" data-driven optimization paradigm that explores the information feedback to fully couple upper-stream machine learning and downstream mathematical programming could be a more effective and rigorous approach.

**4.1.1. The issues of conventional "open-loop" data-driven optimization methods**

It is widely recognized that data-driven optimization is a promising way to hedging against uncertainty in the era of big data and deep learning. Such promise hinges heavily on the organic integration and effective interaction between machine learning and mathematical programming. In existing data-driven optimization frameworks, the tasks performed by the data-driven system and the model-based system are treated separately in a sequential fashion. More specifically, data serve as input to a data-driven system. After that, useful, accurate and relevant uncertainty information is extracted through the data-driven system and further passed along to the model-based system based on mathematical programming for rigorous and systematic optimization under uncertainty, using paradigms such as robust optimization and stochastic programming. However, due to the sequential connection between these two systems, the machine learning model is trained without interacting with the "downstream" mathematical programming. Accordingly, from a control theoretical perspective, such "hybrid" systems in the existing data-driven optimization literature are essentially open loop. In contrast to open-loop systems, closed-loop systems using the feedback control strategy deliver amazingly superior system performance (e.g. stability, robustness to disturbances, and safety) in virtually every area of science and engineering, such as biological systems, social networks, and mechanical systems [191]. Therefore, there should be a "feedback" channel for information flow returning from the model-based system to the data-driven system, in addition to the information flow that is fed into the mathematical programming problem from the machine learning results. The design of such feedback loops from mathematical programming to machine learning deserves further attention in future research. Although there are closed-loop machine learning methods in the case of reinforcement learning [192], to the best of our knowledge, there are few works on developing a closed-loop strategy for data-driven mathematical programming under uncertainty. Different from mathematical programming, reinforcement learning is a kind of machine learning that aims to find an action policy to increase an agent's performance in terms of reward by interacting with an environment.

**4.1.2. A "closed-loop" data-driven optimization framework**

Due to its critical role in the training of machine learning models, loss functions could provide a foundation for considering feedback steps in future research efforts. Instead of using a mathematical-programming-agnostic loss function (e.g. logistic or squared-error loss), a loss function that incorporates the objective function of mathematical programming could be used to train the machine learning model. Specifically, a weighted sum of the conventional loss function and the objective function in the mathematical programming problem should be useful in handling issues experienced with current "open-loop" data-driven frameworks. An iterative scheme between machine learning and mathematical programming offers an alternative promising path to close the loop of the data-driven system and the model-based system. Figure 2 presents the potential schematic of the closed-loop data-driven mathematical programming system. From the figure, we can see that the feedback from the model-based system serves as input to the data-driven system. In this way, the "hybrid" system becomes a closed-loop one in which information can be transmitted in both directions. Such feedback strategy should be beneficial to the "hybrid" system and could provide an effective way to organically integrate machine learning and mathematical programming.

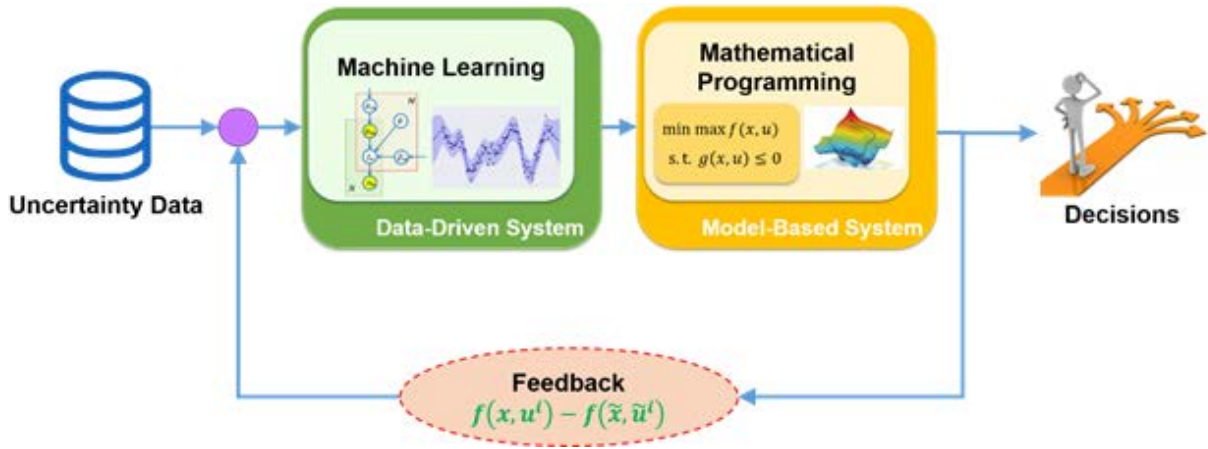

**Figure 2.** The schematic of "closed-loop" data-driven mathematical programming framework.

Research challenges emerge from the feedback step in the "hybrid" system. In typical PSE applications, the problem size of mathematical programs tends to be large. Such large-scale mathematical programming problems in conjunction with big data could pose a computational challenge for the training of machine learning. Additionally, how to design an effective feedback strategy to "close the loop" poses another key challenge to be addressed.

### 4.1.3. Incorporating "prior" knowledge in the data-driven optimization framework

In addition to uncertainty data, some available domain-specific knowledge or "prior" knowledge could serve as another informative input to the data-driven system. Relying solely on the data to develop the uncertainty model could unfavorably influence the downstream mathematical programming. The prior knowledge depicts what the decision maker knows about the uncertainty, and it can come in different forms. For example, the prior knowledge could be the structural information of probability distributions, upper and lower bounds of uncertain parameters or certain correlation relationship among uncertainties. Incorporating such "prior" knowledge in the data-driven optimization framework could be substantially useful and provides more reliable results in the face of messy data.

## 4.2. Leveraging deep learning techniques for hedging against uncertainty in data-driven optimization

Recently, deep learning has shown great promise due to its amazing power in hierarchical representation of data [15]. The deep learning techniques are now shaping and revolutionizing many areas of science and engineering [18]. In recent years, deep learning has a wide array of applications in the PSE domain, such as process monitoring [193, 194], refinery scheduling [195], and soft sensor [196]. For extensive surveys on deep learning in the PSE area, we refer the reader to the review papers on this subject [14, 197]. In real applications, uncertainty data exhibit very complex and highly nonlinear characteristic. Therefore, it should be promising to explore the potential opportunities of leveraging deep neural networks with various architectures to uncover useful patterns of uncertainty data for mathematical programming.

In this section, a variety of deep learning techniques are first summarized along with their unique features from a practical point of view, and future research directions on how to leverage the power of deep learning in optimization under uncertainty are further suggested. Research opportunities of integrating data-driven scenario-based optimization with deep generative models are then presented.

### 4.2.1. Various types of deep learning techniques and their potentials

In this subsection, we present three types of deep learning techniques, including deep belief networks, convolutional neural networks, and recurrent neural networks, and explore their potential applications in data-driven optimization under uncertainty.

- Deep belief networks

Among deep learning techniques, deep belief networks (DBNs) are becoming increasingly popular primarily because its unique feature in capturing a hierarchy of latent features [198]. DBNs essentially belong to probabilistic graphical models and are structured by stacking a series of restricted Boltzmann machines (RBMs). This specific network structure is designed based on the fact that a single RBM with only one hidden layer fall shorts of capturing the intrinsic complexities in high-dimensional data. As the building blocks for DBNs, RBMs are characterized as two layers of neurons, namely hidden layer and visible layer. Note that the hidden layer can be regarded as the abstract representation of the visible layer. There are undirected connections between these two layers, while there exist no intra-connections within each layer. The training process of DBNs typically involves the pre-training and fine-tuning procedures in a layer-wise scheme. Armed with multiple layers of hidden variables, DBNs enjoy unique power in extracting a hierarchy of latent features automatically, which is desirable in many practical applications. As a result, DBNs have been applied in a wide spectrum of areas, including fault diagnosis [194], soft sensor [196], and drug discovery [199]. DBNs can decipher complicated nonlinear correlation among uncertain parameters. Recently, deep Gaussian process model was proposed as a special type of DBN based Gaussian process mappings. Due to its unique advantage in nonlinear regression, deep Gaussian process model should be used to characterize the relationship between uncertain parameters, such as product price and demand.

- Convolutional neural networks

Convolutional neural networks (CNNs) are one specialized version of deep neural networks [200], and they have become increasingly popular in areas such as image classification, speech recognition, and robotics. Inspired by the visual neuroscience, CNNs are designed to fully exploit the three main ideas, namely sparse connectivity, weight sharing, and equivariant representations [15]. This kind of neural network is suited for processing data in the form of multiple arrays, particularly two-dimensional image data. The architecture of a CNN typically consists of convolution layers, nonlinear layers, and pooling layers. In convolution layers, feature maps are extracted by performing convolutions between local patch of data and filters. The filters share the same weights when moving across the dataset, leading to reduced number of parameters in networks. The obtained results are further passed through a nonlinear activation function, such as rectified linear unit (ReLU). After that, pooling layers, such as max pooling and average pooling, are applied to aggregate semantically similar features. Such different types of layers are alternatively connected to extract hierarchical features with various abstractions. For the purpose of classification, a fully connected layer is stacked after extracting the high-level features. Although CNNs are mainly used for image classification, they have been used to learn spatial features of traffic flow data at nearby locations which exhibit strong spatial correlations [201]. Given its unique power in spatial data modeling, CNNs hold the potential to model uncertainty data with large spatial correlations, such as demand data in different adjacent market locations. In addition, the CNNs can be trained for the labeled multi-class uncertainty data to perform the task of classification. Therefore, the output of the CNN potentially acts as the probability weights used in the data-driven stochastic robust optimization framework.

- Recurrent neural networks

Besides the aforementioned models for spatial data, recurrent neural networks (RNNs) are widely recognized as the state-of-the-art deep learning technique for processing time series data, especially those from language and speech [202]. RNNs can be considered as feedforward neural networks if they are unfolded in time scale. The architecture of neural networks in a RNN possesses a unique structure of directed cycles among hidden units. In addition, the inputs of the hidden unit come from both the hidden unit of previous time and the input unit at current time. Accordingly, these hidden units in the architecture of RNNs constitute the state vectors and store the historical information of past input data. With this special architecture, RNNs are well-suited for feature learning for sequential data and demonstrate successful applications in various areas, including natural speech recognition [202], and load forecasting [203]. However, one drawback of RNNs is its weakness in storing long-term memory due to gradient vanishing and exploding problems. To address this issue, research efforts have been made on variants of RNNs, such as long short-term memory (LSTM) and gated recurrent unit (GRU) [204]. By explicitly incorporating input, output and forget gates, LSTM enhances the capability of memorizing the long-term dependency among sequential data. In sequential mathematical programming under uncertainty, massive time series of uncertain parameters are collected. Uncertainty data realized at different time stages often exhibit temporal dynamics. To this end, deep learning techniques, such as deep RNNs and LSTM, could be leveraged to decipher the temporal dynamics and trajectories of uncertainty over time stages. Therefore, exploring the integration between deep learning and multistage optimization under uncertainty is another promising research direction.

**4.2.2. Deep generative models for scenario-based optimization**

Despite the various successful applications of scenario-based optimization, this type of data-driven optimization framework has its own limitations. In general, scenario-based optimization enjoys computational efficiency by constraint sampling and provides the feasibility guarantee regardless of probability types. These advantages of scenario-based optimization rely heavily on the key assumption that sufficient amount of uncertainty data is available. However, in most practical cases, this assumption does not hold, and on the contrary the amount of uncertainty data sampled from the underlying true distribution is quite limited. Moreover, acquiring uncertainty data could be extremely expensive or time consuming in some specific cases, which greatly hinders the applicability of the scenario-based approach [205]. Existing studies of the scenario-based optimization neglect the aforementioned practical situation [166, 169, 182]. The practical challenge of handling insufficient amount of data requires further research attention, and data-driven scenario-based optimization frameworks addressing this issue need to be developed.

This knowledge gap could be potentially filled by leveraging the power of deep generative models for the data-driven scenario-based optimization, whose schematic is shown in Figure 3. Instead of assuming unlimited uncertainty scenarios sampled from the true distribution, deep generative models could be leveraged to learn the intrinsic useful patterns from the available uncertainty data and to generate synthetic data. These synthetic uncertainty data generated by the deep learning techniques mimic the real uncertainty data, and should be potentially useful in the scenario-based optimization model. Deep generative models can be utilized to generate synthetic uncertainty data with the aim for better decision with insufficient uncertainty data. To be more precise, in deep generative models, the true data distribution is learned either explicitly or implicitly, and then the learned distribution is used to generate new data points referred to as synthetic data. One of the most commonly used deep generative models is variational autoencoders (VAEs) [15]. VAEs generate new data samples through the architecture of synthesizing an encoder network and a decoder network in an unsupervised fashion. The function of encoders is to reduce the dimension of input data and extracts the latent features, while the decoder network aims to reconstruct data given the latent variables. In this way, the VAE model learns the complicated target distribution by maximizing the lower bound of the data log-likelihood. The advantage of this technique is that its quality is easily evaluated via log-likelihood or importance sampling. However, researchers have found out that VAEs typically tend to generate blurry images, meaning a noticeable difference between the true distribution and the learned one [15]. Recently, an emerging deep generative model named generative adversarial networks (GANs) was proposed and has become increasingly popular in various areas, such as image processing [206], renewable scenario generation [207], and molecular designs [208]. Different from VAEs, GANs implicitly learn the data distribution through a zero-sum game between two competing neural networks, namely generator network and discriminator network [209]. Given the noise input, the generator network competes against the discriminator network by generating plausible synthetic data. On the contrary, the discriminator network attempts to distinguish the real uncertainty data from the synthetic data. These two networks compete against each other. Accordingly, the data distribution resulted from the generator network will be the true distribution once the Nash equilibrium is achieved. The required sample size for random convex programs scales linearly with the number of decision variables [181], implying that the "small data" regime should be frequently encountered for large-scale optimization problems. Consequently, data-driven scenario-based optimization tends to suffer severely from the issue of insufficient uncertainty data. Leveraging the power of deep generative models could be a promising way to addressing this challenge. The required sample size to guarantee the constraint satisfaction could become large for optimization problems with a huge number of decision variables and a small value of risk level [181]. Therefore, the available amount of uncertainty data might not be enough for the purpose of probabilistic guarantee. However, the number of uncertainty data can still be sufficient for training generative models.

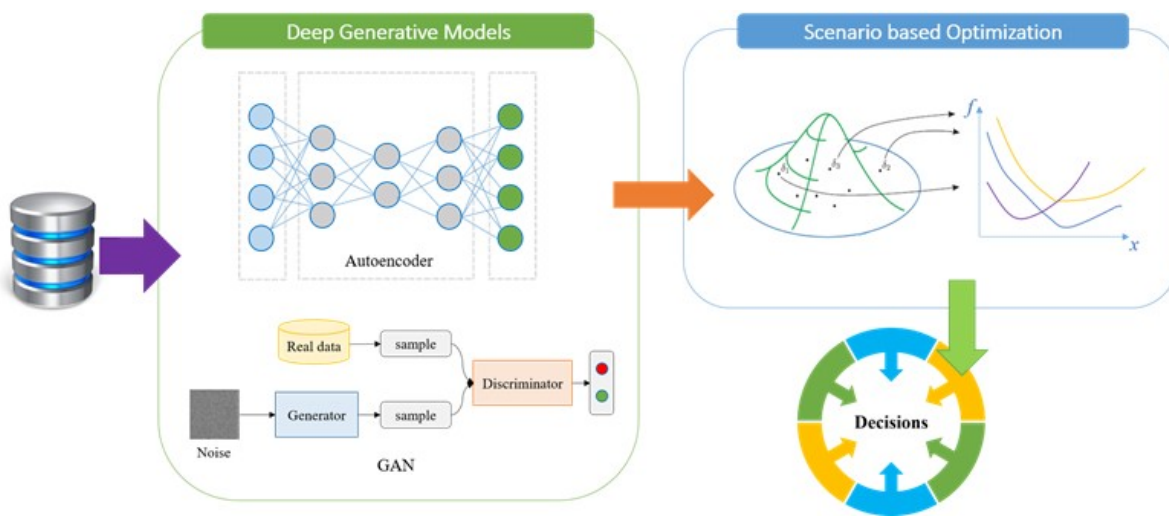

**Figure 3.** The schematic of the scenario-based optimization framework based on deep learning.

Although deep learning could be a silver bullet in many areas, a lot of research challenges still persist in organically integrating the state-of-the-art deep learning techniques with optimization under uncertainty. The discussion in Section 4.2 is aimed to serve as a good starting point to promote the employment of deep learning in the field of data-driven optimization under uncertainty.

## 4.3. Online learning-based data-driven optimization: a learning-while-optimizing paradigm for addressing uncertainty

In conventional data-driven optimization frameworks, a batch of uncertainty data serves as input to the data-driven system, in which learning typically takes place only once and is termed as batch machine learning. Most, if not all, of the papers on data-driven optimization under uncertainty are restricted to such learning of data [25, 116, 144, 157], so they fail to account for real-time data. For example, in data-driven robust optimization methods, uncertainty sets are learned from a batch of uncertainty data. Once these data-driven uncertainty sets are obtained, they remain fixed for the model-based system based on mathematical programming and are not updated or refined. Additionally, probability distributions of uncertainties and their support sets could be time variant and evolves gradually, rendering the data-driven system "outdated". Such obsolete data-driven system inevitably deteriorates the resulting solution quality of the mathematical programming problem. In many practical settings, uncertainty data are collected sequentially in an online fashion [210]. Although previous works have explored the online learning of uncertainty sets [162], they typically re-train the data-driven system from scratch using the existing and new addition of data, thus making these approaches suitable only for systems with slow dynamics. Therefore, few studies to date investigate the real-time data analytics for systems with fast dynamics such as those encountered in chemical processes, establishing research opportunities for the PSE community.

An online-learning-based data-driven optimization paradigm, in which learning takes place iteratively to account for real-time data, could be a promising research direction. More specifically, a learning-while-optimizing scheme could be explored by taking advantage of deep reinforcement learning. On one hand, the uncertainty model should be time varying to accommodate real-time uncertainty data. On the other hand, decisions are made sequentially under uncertainty. After decisions are made, uncertainties are realized and then collected in the database. There are research challenges associated with such online-learning-based frameworks. Updating the data-driven system in an online fashion is paramount in implementing the learning-while-optimizing scheme and poses a key research challenge. Additionally, developing efficient algorithms to solve the resulting online-learning-based mathematical programming problems creates the computational challenge. There exist some theoretical research challenges as well. One theoretical challenge is to investigate the convergence of solutions when the probability distribution shift to a new one. Another challenge is to provide theoretical bounds for computational complexity and required memory for the online-learning-based data-driven optimization.

## 5. Conclusions

Although conventional stochastic programming, robust optimization, and chance constrained optimization are the most recognized modeling paradigms for hedging against uncertainty, it is foreseeable that in the near future data-driven mathematical programming frameworks would experience a rapid growth fueled by big data and deep learning. We reviewed recent progress of data-driven mathematical programming under uncertainty in terms of systematic uncertainty modeling, organic integration of machine learning and mathematical programming, and efficient computational algorithms for solving the resulting mathematical programming problems. The advantages and disadvantages of different data-driven uncertainty models were also analyzed in detail. Future research could be directed toward devising feedback steps to close the loop of the data-driven system and the model-based system, leveraging the power of deep generative models for the data-driven scenario-based optimization, and developing data-driven mathematical programming frameworks with online learning for real-time data.

[179] J. A. De Loera, R. N. La Haye, D. Oliveros, and E. Roldan-Pensado, "Chance-Constrained Convex Mixed-Integer Optimization and Beyond: Two Sampling Algorithms within S-Optimization," *Journal of Convex Analysis,* vol. 25, pp. 201-218, 2018.

[180] M. Chamanbaz, F. Dabbene, R. Tempo, V. Venkataramanan, and Q. G. Wang, "Sequential Randomized Algorithms for Convex Optimization in the Presence of Uncertainty," *IEEE Transactions on Automatic Control,* vol. 61, pp. 2565-2571, 2016.

[181] T. Alamo, R. Tempo, A. Luque, and D. R. Ramirez, "Randomized methods for design of uncertain systems: Sample complexity and sequential algorithms," *Automatica,* vol. 52, pp. 160-172, 2015.

[182] G. Calafiore, "Repetitive Scenario Design," *IEEE Transactions on Automatic Control,* vol. 62, pp. 1125-1137, 2017.

[183] K. Margellos, A. Falsone, S. Garatti, and M. Prandini, "Distributed Constrained Optimization and Consensus in Uncertain Networks via Proximal Minimization," *IEEE Transactions on Automatic Control,* vol. 63, pp. 1372-1387, 2018.

[184] L. Carlone, V. Srivastava, F. Bullo, and G. C. Calafiore, "Distributed Random Convex Programming via Constraints Consensus," *SIAM Journal on Control and Optimization,* vol. 52, pp. 629-662, 2014.

[185] K. You, R. Tempo, and P. Xie, "Distributed Algorithms for Robust Convex Optimization via the Scenario Approach," *IEEE Transactions on Automatic Control,* pp. 1-1, 2018.

[186] A. Care, S. Garatti, and M. C. Campi, "FAST-Fast Algorithm for the Scenario Technique," *Operations Research,* vol. 62, pp. 662-671, 2014.

[187] M. C. Campi, S. Garatti, and F. A. Ramponi, "A General Scenario Theory for Nonconvex Optimization and Decision Making," *IEEE Transactions on Automatic Control,* vol. 63, pp. 4067-4078, 2018.

[188] T. Alamo, R. Tempo, and E. F. Camacho, "Randomized Strategies for Probabilistic Solutions of Uncertain Feasibility and Optimization Problems," *IEEE Transactions on Automatic Control,* vol. 54, pp. 2545-2559, 2009.

[189] G. Calafiore, F. Dabbene, and R. Tempo, "Research on probabilistic methods for control system design," *Automatica,* vol. 47, pp. 1279-1293, 2011.

[190] S. Grammatico, X. J. Zhang, K. Margellos, P. Goulart, and J. Lygeros, "A Scenario Approach for Non-Convex Control Design," *IEEE Transactions on Automatic Control,* vol. 61, pp. 334-345, 2016.

[191] K. J. Åström and P. R. Kumar, "Control: A perspective," *Automatica,* vol. 50, pp. 3-43, 2014.

[192] J. Shin and J. H. Lee, "Multi-timescale, multi-period decision-making model development by combining reinforcement learning and mathematical programming," *Computers & Chemical Engineering,* vol. 121, pp. 556-573, 2019.

[193] W. Zhu, Y. Ma, M. G. Benton, J. A. Romagnoli, and Y. Zhan, "Deep learning for pyrolysis reactor monitoring: From thermal imaging toward smart monitoring system," *AIChE Journal,* vol. 65, pp. 582-591, 2019.

[194] Z. P. Zhang and J. S. Zhao, "A deep belief network based fault diagnosis model for complex chemical processes," *Computers & Chemical Engineering,* vol. 107, pp. 395-407, 2017.

[195] X. Y. Gao, C. Shang, Y. H. Jiang, D. X. Huang, and T. Chen, "Refinery scheduling with varying crude: A deep belief network classification and multimodel approach," *AIChE Journal,* vol. 60, pp. 2525-2532, 2014.